\documentclass[a4paper]{cas-dc}

\usepackage[square,numbers,sort&compress]{natbib}
\usepackage{graphicx}%
\usepackage{multirow}%
\usepackage{amsmath,amssymb,amsfonts}%
\usepackage{amsthm}%
\usepackage{mathrsfs}%
\usepackage[title]{appendix}%
\usepackage{xcolor}%
\usepackage{textcomp}%
\usepackage{manyfoot}%
\usepackage{booktabs}%
\usepackage[ruled,vlined]{algorithm2e}
\usepackage{bm}
\usepackage{algpseudocode}%
\usepackage{listings}%
\usepackage{ulem}
\usepackage{tabularx}
\usepackage{hhline}
\usepackage{caption}
\usepackage{bbding}
\usepackage{pifont}
\usepackage{utfsym}
\usepackage{fontawesome}
\def\tsc#1{\csdef{#1}{\textsc{\lowercase{#1}}\xspace}}
\tsc{WGM}
\tsc{QE}
\begin{document}
\let\WriteBookmarks\relax
\def\floatpagepagefraction{1}
\def\textpagefraction{.001}

\shorttitle{Discrepancy-based Diffusion Models for Lesion Detection in Brain
MRI}    

\shortauthors{K.Fan et. al.}  

\title [mode = title]{Discrepancy-based Diffusion Models for Lesion Detection in Brain MRI}  

% \tnotemark[<tnote number>] 


\author[1]{Keqiang Fan}[
% type=editor,
                        % auid=000,bioid=1,
                        style=chinese,
                        % prefix=Sir,
                        % role=Researcher,
                        orcid=0000-0002-9411-2892
                        ]

\cormark[1]


\ead{kf1d20@soton.ac.uk}

% \ead[url]{www.cvr.cc, cvr@sayahna.org}

% \credit{Conceptualization of this study, Methodology, Software}

\affiliation[1]{organization={Electronics and Computer Science, University of Southampton},
    % addressline={Radarweg 29}, 
    city={ Southampton},
    % citysep={}, % Uncomment if no comma needed between city and postcode
    postcode={SO17 1BJ}, 
    country={UK}}

\author[1]{Xiaohao Cai}[style=chinese,
                        orcid=0000-0003-0924-2834]
\ead{x.cai@soton.ac.uk}
\author[1]{Mahesan Niranjan}[%
    style=chinese,
    orcid=0000-0001-7021-140X]
   % suffix=Jr,
   % ]
\ead{mn@ecs.soton.ac.uk}

\cortext[1]{Corresponding author}



\begin{abstract}
Diffusion probabilistic models (DPMs) have exhibited significant effectiveness in computer vision tasks, particularly in image generation. However, their notable performance heavily relies on labelled datasets, which limits their application in medical images due to the associated high-cost annotations.
Current DPM-related methods for lesion detection in medical imaging, which can be categorized into two distinct approaches, primarily rely on image-level annotations.
The first approach, based on anomaly detection, involves learning reference healthy brain representations and identifying anomalies based on the difference in inference results.
In contrast, the second approach, resembling a segmentation task, employs only the original brain multi-modalities as prior information for generating pixel-level annotations. In this paper, our proposed model -- discrepancy distribution medical diffusion ({DDMD}) -- for lesion detection in brain MRI introduces a novel framework by incorporating distinctive discrepancy features, deviating from the conventional direct reliance on image-level annotations or the original brain modalities. In our method, the inconsistency in image-level annotations is translated into distribution discrepancies among heterogeneous samples while preserving information within homogeneous samples.
This property retains pixel-wise uncertainty and facilitates an implicit ensemble of segmentation, ultimately enhancing the overall detection performance. Thorough experiments conducted on the BRATS2020 benchmark dataset containing multimodal MRI scans for brain tumour detection demonstrate the great performance of our approach in comparison to state-of-the-art methods.
\end{abstract}


% Keywords
\begin{keywords}
Diffusion probabilistic model  \sep
anomaly detection  \sep
segmentation  \sep
brain MRI 
 
\end{keywords}

\maketitle

%----------
\section{Introduction}
%----------
Medical image analysis holds a crucial position in clinical therapy, owing to the significance of digital medical imaging in contemporary healthcare \cite{de2016machine}.
Given the diverse range of data modalities in medical imaging, reliance solely on medical data with image-level annotations is often inadequate for obtaining meaningful diagnostic regions. 
To overcome this constraint, automated and dependable analysis systems, commencing at the pixel level, possess the capability of interpreting medical images, thereby enhancing diagnostic efficiency.
Over the last decades, significant effort has been devoted to developing support tools for assisting radiologists in evaluating medical images \cite{kawamoto2005improving}. These tools include widely used convolutional neural networks (CNNs) \cite{li2021survey} and vision transformers \cite{han2022survey}.
While these predominantly supervised methods hold promise in diminishing the time and effort needed for pixel-level annotation and enhancing result accuracy and consistency, they face challenges with diverse modalities in medical imaging. Moreover, these challenges also arise from limitations inherent in supervised methods, particularly the necessity for extensive expert-annotated data.

Understanding patterns and features within medical data, specifically transitioning from the image-level to pixel-level while simultaneously reducing the demand for annotations, is particularly crucial. 
This challenge has found alleviation in the domain of unsupervised anomaly detection (UAD). 
In UAD, where no supervision discrepancy signals indicating normal or anomalous samples are provided during training, models typically concentrate on reconstructing images from a healthy training distribution \cite{fernando2021deep}. 
Autoencoder (AE) \cite{ballard1987modular}, the widely used architecture, serves as a prevalent model in this domain \cite{sato2018primitive,wang2016research}. 
Trained models can reconstruct inputs from the learned compressed representations between the encoder and decoder. 
Variational autoencoder (VAE) \cite{kingma2013auto}, a variant of AE, assumes that the compressed representation is sampled from a probability distribution, aiming to estimate the parameters of that distribution. The relevant work is proposed to for example detect skin abnormalities in \cite{lu2018anomaly}. Another class of AEs is adversarial AEs, more widely known as generative adversarial network (GAN) \cite{goodfellow2014generative}. 
GANs incorporate their inherent architecture into anomaly detection by employing a generator to produce instances mirroring the normal data encountered during training, while the discriminator serves a pivotal role in distinguishing anomalies by identifying instances \cite{schlegl2019f}.

Recently, diffusion probabilistic models (DPMs) \cite{ho2020denoising} have demonstrated remarkable success in image synthesis \cite{dhariwal2021diffusion,nichol2021improved}, surpassing other models such as GANs or AEs. Within the medical domain, numerous datasets encounter substantial class imbalance attributed to the infrequent occurrence of certain pathologies. Diffusion models alleviate this restriction by generating a variety of images, thereby addressing the limitations imposed by class imbalance in the medical field.
Considering the conditions imposed by image-level annotations, the integration with diffusion models has emerged as a prevalent technique to generate pixel-level annotations, leading to significant achievements in medical image segmentation \cite{xing2023diff,wu2022medsegdiff}.
Despite their success, the straightforward application of a diffusion model with image-level annotations is insufficient.
Owing to the inherent characteristics of medical data, normal and abnormal samples corresponding to the same pathology display a notable degree of similarity. In anomaly detection, it is common for the lesion area to occupy only a fraction of the entire image, which is significantly smaller than other normal regions.

In this paper, to address the above-mentioned issues/challenges, we propose a novel method called discrepancy distribution medical diffusion ({DDMD}) for lesion detection in medical imaging like brain MRI. It also has the potential to serve as a more general approach for anomalous areas detection in image analysis. DDMD integrates additional information extracted from image-level annotations into the DPM for achieving high-quality pixel-level brain MRI annotations.
We utilize image-level labels in the image reconstruction process to draw internal and external differences between different categories. The distribution disparities between heterogeneous samples and the information within homogeneous samples are prominently maintained by our method throughout the inference process of the DPM.  Our contributions are summarized as follows:
\begin{itemize}
    \item We propose a novel DPM-based model DDMD for medical lesion detection in brain MRI. We demonstrate the insufficiency of solely relying on image-level annotations and the importance of the distribution discrepancies preservation in medical lesion detection performance enhancement.

    \item We propose a discrepancy guidance strategy to direct the denoising procedure in our model, conditioning each step with inter-discrepancy and intra-discrepancy in the diffusion process.

    \item We conduct thorough quantitative and qualitative experiments, demonstrating the superior performance of our proposed method in anomaly detection and segmentation tasks in brain MRI in comparison to the state-of-the-art methods.
\end{itemize}

The rest of the paper is organized as follows. In Section \ref{sec:relatedwork}, we briefly review the related medical anomaly detection methods and diffusion probabilistic models. Section \ref{sec:method} covers the background knowledge of diffusion models and introduces the proposed DDMD method. Sections \ref{sec:Experimental Setup} and \ref{sec:Results and Discussion} present the details of the experimental settings and the corresponding results, respectively. Finally, Section \ref{sec:conclusion} concludes and points to potential future works.

%---------
\section{Related Work}\label{sec:relatedwork}
%---------

In this section, we briefly recall the related work regarding medical anomaly detection and the use of DPMs for lesion detection in medical imaging.

%---------
\subsection{Medical Anomaly Detection}
%---------
In medical domain, the relationship between anomaly detection and segmentation techniques is interrelated, contributing to early disease detection and localization.
In medical image segmentation, reconstruction-based networks such as U-Net \cite{ronneberger2015u} and SegNet \cite{badrinarayanan2017segnet} are common methods to predict segmentation masks for  input images and have been widely applied to various tasks \cite{li2018h,xiao2020segmentation}. 
Some foundational frameworks for medical anomaly detection originated from the utilization of AEs \cite{ballard1987modular} or VAEs \cite{kingma2013auto}. These models are trained on healthy data, learning to identify anomalous areas through the reconstruction difference between healthy and anomalous inputs \cite{sato2018primitive,wang2016research,lu2018anomaly}. However, these methodologies exhibit inherent shortcomings. For example, AE-based models have demonstrated reliable training and fast inference, but they encounter difficulties in effectively modelling high-dimensional data distributions, resulting in imprecise and blurry reconstructions \cite{baur2021autoencoders}. 
Alternative approaches like GANs, leveraging synthetic anomaly images during training, adopt discriminative strategies to detect and localize anomalies during reconstruction \cite{ristea2022self, zavrtanik2021draem}.
Nonetheless, the training of GANs presents challenges like demanding extensive hyperparameter tuning. Moreover, GAN-based methods encounter difficulties such as under-representation of crucial features, as noted in \cite{kazeminia2020gans}, and demonstrate suboptimal performance on datasets characterized by class imbalance.

\subsection{Diffusion Probabilistic Models}

Recently, DPMs have garnered attention for their capacity to outperform GANs in image synthesis \cite{dhariwal2021diffusion}.
Diverging from other methods, DPMs learn complex data distributions through the integration of both forward and reverse diffusion processes.
During the denoising process, the original data structure is recovered from the perturbed data distribution, transitioning from the pure Gaussian distribution induced by the forward diffusion process.
The flexibility and tractability of this process significantly enhance its efficacy in image generation, thereby instigating extensive research into its potential application for lesion detection in medical images \cite{wyatt2022anoddpm,wolleb2022diffusion,sanchez2022healthy,pinaya2022fast,behrendt2023patched,wolleb2022diffusion_seg,wu2023medsegdiffv2,wu2022medsegdiff}.
Concerning lesion detection within existing DPM-related methods, the majority rely heavily on image-level annotations. 
Specifically, these methods involve establishing a reference representation of a healthy brain and identifying anomalies by discerning differences in inference outcomes \cite{wyatt2022anoddpm,wolleb2022diffusion,sanchez2022healthy,pinaya2022fast,behrendt2023patched}. 
An alternative strategy involves directly leveraging the original brain multi-modalities as fundamental information for generating pixel-level annotations \cite{wolleb2022diffusion_seg,wu2023medsegdiffv2,wu2022medsegdiff}.
These models utilize a stochastic sampling process to generate an implicit representation of pixel-level annotations, resulting in improved final segmentation performance.

%----------
\section{Method}\label{sec:method}
%----------
\begin{figure*}[t]
\vspace{0.4cm}
\centering
\includegraphics[width=0.96\textwidth, height=0.65\textheight]{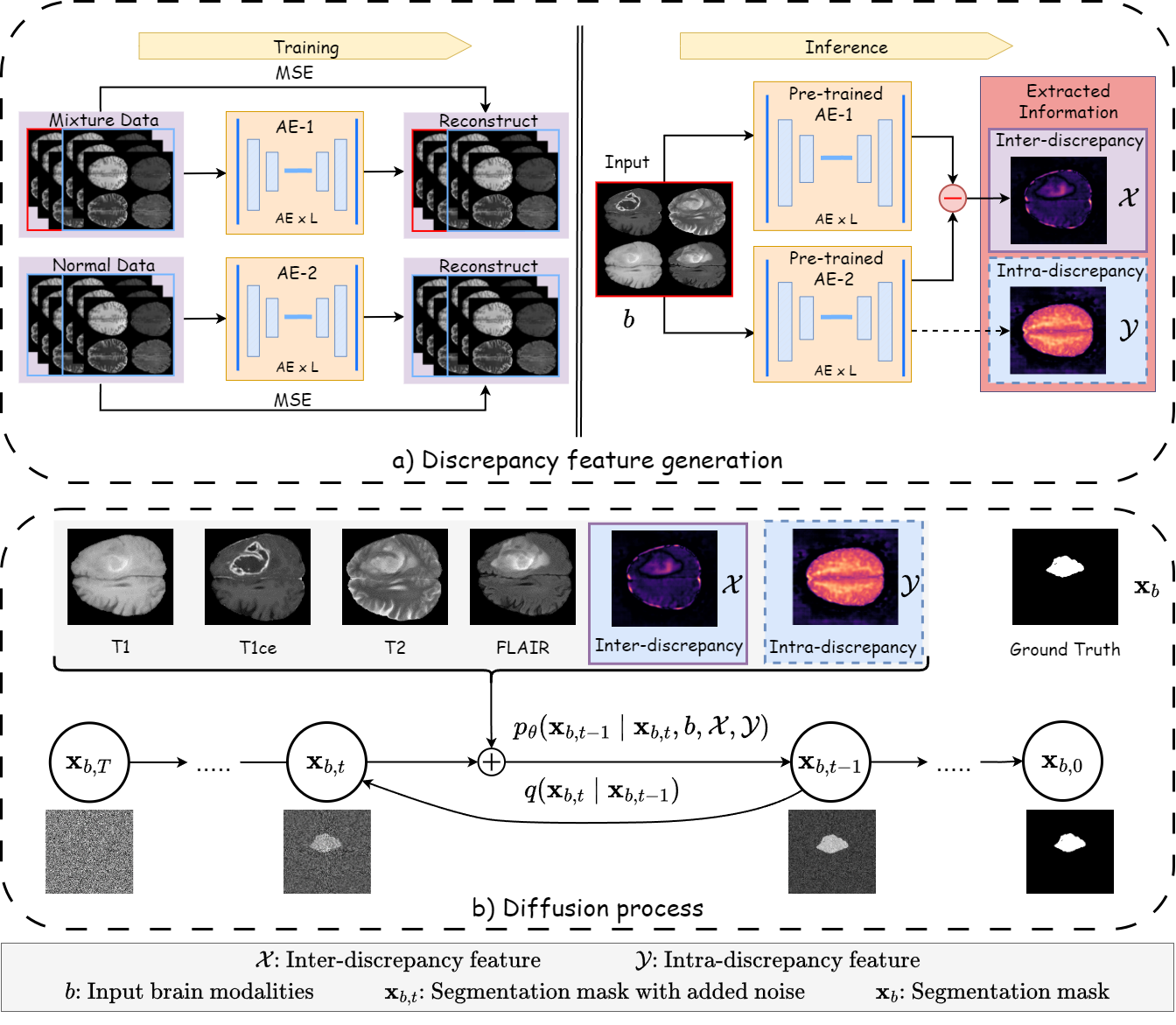}
\caption{ The framework of the proposed {DDMD} method. It consists of two key processes: a) the discrepancy feature generation process and b) the diffusion process.  }
\label{fig:framework}
\vspace{0.2cm}
\end{figure*}

We now present our proposed {DDMD} approach, a framework designed to ascertain the localization of brain lesion areas given a set of brain modalities.
The DDMD effectively governs the implicit diffusion process, utilizing the effective discrepancy features extracted from the image-level annotation to generate pixel-level annotations.
Inspired by the methodology outlined in DPM \cite{ho2020denoising}, our model comprises two fundamental processes: the discrepancy feature generation process and the diffusion process. In generating discrepancy features, we leverage image-level annotations associated with the input brain modalities to derive pixel-level discrepancies through the reconstruction process.
Subsequently, in the diffusion process, a DPM model is trained using these discrepancy features as prior information in conjunction with the input brain modalities, ultimately generating a segmentation mask that identifies the corresponding brain lesion area. The architecture detail of {DDMD} is illustrated in Fig. \ref{fig:framework}.

\subsection{Background on Diffusion Models}
\begin{figure}[tb]
    \centering
    \includegraphics[width=3.35in,height=1.0in]{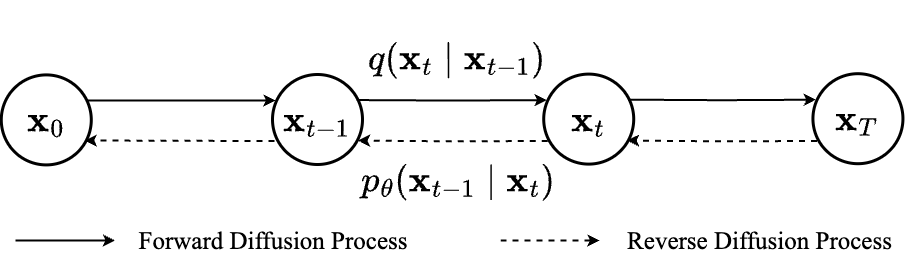}
    \caption{Forward and reverse diffusion processes in DPM.}
    \label{fig:diffusion_process}
\end{figure}

The background on diffusion models served as the preliminary of our DDMD method is firstly introduced below. Diffusion models are a powerful class of probabilistic generative models that are used to learn complex data distributions. As shown in Fig. \ref{fig:diffusion_process}, these models accomplish the learning task by utilizing two key stages: the forward diffusion process and the reverse diffusion process. 

The forward diffusion process systematically induces perturbations in input data distribution through a gradual escalation of noise levels, ultimately resulting in the transformation of the data into pure Gaussian noise.
Let $\mathbf{x}_0$ represent an original image, and denote the original distribution of $\mathbf{x}_0$ as $q(\mathbf{x}_0)$. Consequently, we have $\mathbf{x}_0 \sim q(\mathbf{x}_0)$. As the noising step $t \in \{0, 1, \ldots, T\}$ increases, a sequence of noisy images, $\mathbf{x}_1, \mathbf{x}_2, \cdots, \mathbf{x}_T$, at each diffusion step can be computed by progressively introducing Gaussian noise to $\mathbf{x}_0$ through a Markovian process:
% \vspace{-1.5em}
\begin{align}
q(\mathbf{x}_t|\mathbf{x}_{t-1}) & = \mathcal{N}(\mathbf{x}_t; \sqrt{1-\beta_t} \cdot \mathbf{x}_{t-1}, \beta_t \cdot \mathbf{I}), 
% ,\forall t\in\{1,\ldots,T\},
\label{eq:Markovian process} \\
q(\mathbf{x}_{1:T}|\mathbf{x}_0) & = \prod_{t=1}^T q(\mathbf{x}_t|\mathbf{x}_{t-1}),
\end{align}
where $T$ represents the total number of noising steps, $\beta_t\in[0,1)$ is the hyper-parameter for the variance schedule across diffusion steps, $\mathbf{I}$ is the identity matrix, and $\mathcal{N}(\mathbf{x};\mu,\sigma)$ represents the normal distribution with mean $\mu$ and covariance $\sigma$. 
Let $\alpha_t=1-\beta_t$ and $\bar{\alpha}_t=\prod_{i=1}^t\alpha_i$, we have
\begin{align}
    q(\mathbf{x}_t|\mathbf{x}_0) & =\mathcal{N}(\mathbf{x}_t;\sqrt{\bar{\alpha}_t}\cdot \mathbf{x}_0,(1-\bar{\alpha}_t)\cdot\mathbf{I}),
\label{eq:Markovian process_t_2_0}    \\
\mathbf{x}_{t}& =\sqrt{\bar{\alpha}_{t}}\mathbf{x}_{0}+\sqrt{1-\overline{\alpha}_{t}}\epsilon,
\label{eq:Markovian process_formula}
\end{align} 
where $\epsilon$ represents the Gaussian noise. As $T \rightarrow \infty$, $\mathbf{x}_T$ converges to pure Gaussian noise.

The reverse diffusion process is employed to reconstruct the original image $\mathbf{x}_0$ from pure Gaussian noise 
% $\mathbf{x}_t = \epsilon \sim \mathcal{N}(0, \mathbf{I})$ 
$\mathbf{x}_T~\sim~\mathcal{N}(0, \mathbf{I})$ 
from $T \rightarrow 0$ through the reverse Markov chain. According to the work in \cite{ho2020denoising}, the reverse diffusion process say $p_\theta$ can be learned from the 
network $\theta$, i.e.,
\begin{align}
\mathbf{x}_0 \sim p_{\theta}\left(\mathbf{x}_{0:T}\right) & = p_{\theta}\left(\mathbf{x}_{T}\right) \prod_{t=1}^{T} p_{\theta}\left(\mathbf{x}_{t-1} \mid \mathbf{x}_{t}\right),
\label{eq:the reverse diffusion process_formula} \\
p_{\theta}\left(\mathbf{x}_{t-1} \mid \mathbf{x}_{t}\right) & = \mathcal{N}\left(\mathbf{x}_{t-1}; \mu_{\theta}\left(\mathbf{x}_{t}, t\right), \Sigma_{\theta}\left(\mathbf{x}_{t}, t\right)\right),
\label{eq:the reverse diffusion process_distribution}
\end{align} 
where $\mu_\theta$ and $\Sigma_\theta$ are parameters estimated by the 
% predicted noise $\epsilon_\theta$ from the 
network $\theta$. The objective is to utilize the predicted distribution $p_{\theta}(\mathbf{x}_0)$ to approximate the true data distribution $q(\mathbf{x}_0)$ and recover the original image. As shown in \cite{ho2020denoising}, the original loss function, the variational lower bound (VLB), can be simplified and parameterized into 
\begin{equation} L_{\text{simple}}=E_{t,\mathbf{x}_0,\epsilon}\left[\left\|\epsilon-\epsilon_\theta\left(\mathbf{x}_t,t\right)\right\|^2\right],
\label{eq:original_loss_function}
\end{equation} 
where $\epsilon_\theta(\mathbf{x}_t, t)$ is a noise model. Instead of directly parameterizing $\mu_\theta(\mathbf{x}_t, t)$ as a neural network, a noise model 
$\epsilon_\theta(\mathbf{x}_t, t)$ is trained  to approximate the ground-truth noise $\epsilon$. Following \cite{ho2020denoising}, a U-net model is used to execute a series of step-by-step denoising operations with learned parameters, resulting in the restoration of the original image.
Starting from a Gaussian noise $\mathcal{N}(0, \mathbf{I})$, the inference result $\mathbf{x}_{t-1}$ can then be predicted from $\mathbf{x}_{t}$ iteratively, i.e.,   
\begin{equation}
    \mathbf{x}_{t-1}=\frac{1}{\sqrt{\alpha_{t}}}\left(\mathbf{x}_{t}-\frac{1-\alpha_{t}}{\sqrt{1-\overline{\alpha}_{t}}}\epsilon_{\theta}(\mathbf{x}_{t},t)\right)+\sigma_{t}\mathbf{z},
\label{eq:inference result}
\end{equation} where $\mathbf{z}\sim\mathcal{N}(0,\mathbf{I})$, $t=T,...,1$ and
$\sigma_t$ is the variance schema that can be learned by the model, as proposed in \cite{nichol2021improved}.

\subsection{Discrepancy Feature Generation}\label{Discrepanvy_feature_Generation}
We use image-level labels to obtain pixel-level disparities inherent in the input data systematically. The features obtained through this process encapsulate crucial information regarding class disparities in the input images, which will be used to enhance the effectiveness of the subsequent segmentation and detection tasks.

With abuse of notation, let $\mathbf{b}$ also represent the given image, composed of $C$ modalities, each with size of $1 \times H \times W$. Thus, the size of $\mathbf{b}$ is $C \times H \times W$, where $C$, $H$ and $W$ denote the number of modalities, the image height and width, respectively. Additionally, let $\mathbf{x}_b$ denote the corresponding segmentation mask of $\mathbf{b}$ with size of $1 \times H \times W$ As previously mentioned, our approach involves utilizing image-level annotations to generate discrepancy features.
In consideration of tumour presence within brain MRI images, we introduce $y \in \{0, 1\}$ from the segmentation mask $\mathbf{x}_b$ as its image label, where the values $0$ and $1$ signify the image is normal and abnormal, respectively.
Then, two datasets $\mathcal{I}_A$ and $\mathcal{I}_B$ can be formed based on the image labels, where $\mathcal{I}_A = \{\mathbf{b}_{Ai}, i=1, \ldots, N\}$ containing $N$ normal images and $\mathcal{I}_B = \{\mathbf{b}_{Bi}, i=1, \ldots, M\}$ containing $M$ images which can be either normal or abnormal. Therefore, in our experiments, $\mathcal{I}_A$ and $\mathcal{I}_B$ represent the normal dataset and the mixture dataset, respectively.

As shown in the training stage of the discrepancy feature generation process in Fig.~\ref{fig:framework}-a, we employ two modules, denoted as $\text{AE-1}$ and $\text{AE-2}$, to learn the distribution of the disparate datasets. 
Each module comprises $L$ number of AE networks with identical architectures but different parameter initializations, which can enhance representation learning and increase adaptability to variations in data distribution.
We use the mean squared error (MSE) loss to train these networks, aiming to minimize reconstruction errors. Specifically, $\text{AE-1}$ is trained on the mixture dataset $\mathcal{I}_B$ with
% to learn the mixture data distribution:
\begin{equation}
\centering
    \mathcal{L}_{\text{AE-1}}=\frac1{M}\sum_{\mathbf{b}_B\in \mathcal{I}_B}\sum_{j=1}^L\|\mathbf{b}_B-\mathbf{\hat{b}}_{Bj}\|^2,
\end{equation} 
and $\text{AE-2}$  is trained on the normal dataset $\mathcal{I}_A$ with
% to learn the normal data distribution:
\begin{equation}
    \mathcal{L}_{\text{AE-2}}=\frac1{N}\sum_{\mathbf{b}_A\in \mathcal{I}_A}\sum_{j=1}^L\|\mathbf{b}_A-\mathbf{\hat{b}}_{Aj}\|^2,
\end{equation} 
where $\mathbf{\hat{b}}_{Aj}$ and $\mathbf{\hat{b}}_{Bj}$ (each with size of $C \times H \times W$) are the reconstructed results of $\mathbf{b}_A \in \mathcal{I}_A$ and $\mathbf{b}_B \in \mathcal{I}_B$ using the $j$-th AE model, respectively.
Through these two loss functions, $\text{AE-1}$ is designed to learn the mixed brain data distribution and acquire information across all classes, while $\text{AE-2}$ is specifically trained to capture the normal brain data distribution. 
Therefore, when both $\text{AE-1}$ and $\text{AE-2}$ take abnormal data as input, the dissimilarity in their reconstruction results indicates the abnormal class discrepancy. On the other hand, the absence of discrepancies indicates that the input data corresponds to  normal brain images. 
%-----
%-----

As shown in Fig. \ref{fig:framework} the inference stage of the discrepancy feature generation process, given a test image $\mathbf{b}$  with $C$ modalities as input, we first calculate the average reconstruction results from the AE models for modules $\text{AE-1}$ and $\text{AE-2}$, denoted as $\hat{\bm{\mu}}^{\text{AE-1}}_b$ and $\hat{\bm{\mu}}^{\text{AE-2}}_b$, respectively, i.e., 
\begin{equation}
\hat{\bm{\mu}}^{\text{AE-1}}_b=\frac1L\sum_{j=1}^L\hat{\mathbf{b}}_{j}^{\text{AE-1}}, \ \
\hat{\bm{\mu}}^{\text{AE-2}}_b=\frac1L\sum_{j=1}^L\hat{\mathbf{b}}_{j}^{\text{AE-2}},
\end{equation} where 
$\hat{\mathbf{b}}_{j}^{\text{AE-1}}$ and $\hat{\mathbf{b}}_{j}^{\text{AE-2}}$ (each with size of $C \times H \times W$) are the reconstruction results of the test image $\mathbf{b}$ from the $j$-th AE model in modules AE-1 and AE-2, respectively.
The size of $\hat{\bm{\mu}}^{\text{AE-1}}_b$ and $\hat{\bm{\mu}}^{\text{AE-2}}_b$ are therefore both $C \times H \times W$. Then, the inter-discrepancy features and the intra-discrepancy features, denoted respectively as $\mathcal{X}$ and $\mathcal{Y}$ both with size of $H \times W$, are generated by averaging across all the $C$ modalities from $\hat{\bm{\mu}}^{\text{AE-1}}_b$, $\hat{\bm{\mu}}^{\text{AE-2}}_b$ and $\hat{\mathbf{b}}_{j}^{\text{AE-2}}$, i.e.,
\begin{align} 
\mathcal{X}& =\frac1C\sum_{c=1}^C|\hat{\bm{\mu}}^{\text{AE-1}}_{bc} - \hat{\bm{\mu}}^{\text{AE-2}}_{bc}|,  \label{eqn:inter-dis-x}
\\
\mathcal{Y}& =\sqrt{\frac1L\frac1C\sum_{j=1}^L\sum_{c=1}^C(\hat{\mathbf{b}}_{jc}^{\text{AE-2}} - \hat{
\bm{\mu}}^{\text{AE-2}}_{bc})^2}, \label{eqn:intra-dis-y}
\end{align} 
where $\hat{\bm{\mu}}^{\text{AE-1}}_{bc}$, $\hat{\bm{\mu}}^{\text{AE-2}}_{bc}$ and $\hat{\mathbf{b}}_{jc}^{\text{AE-2}}$ (each with size of $H \times W$) are the $c$-th modality of $\hat{\bm{\mu}}^{\text{AE-1}}_b$, $\hat{\bm{\mu}}^{\text{AE-2}}_b$ and $\hat{\mathbf{b}}_{j}^{\text{AE-2}}$, respectively; note that point-wise operations are used above to calculate $\mathcal{X}$ and $\mathcal{Y}$, therefore leading to the size of $H \times W$.  
Moreover, the mean values of $\mathcal{X}$ and $\mathcal{Y}$, named respectively the inter-discrepancy score and intra-discrepancy score for the whole data modalities of the test image $\mathbf{b}$, can be utilized as indicators of the discrepancy degree. The inter-discrepancy score and intra-discrepancy score for each data modality of the test image $\mathbf{b}$ can be calculated in the same manner by just considering the $c$-th modality in Eq. \eqref{eqn:inter-dis-x} and \eqref{eqn:intra-dis-y}, where $c = 1, 2, \ldots, C$.

%----------
\subsection{Diffusion with Discrepancy Features}
%----------
As aforementioned, we draw inspiration from the methodology outlined in DPM \cite{ho2020denoising, nichol2021improved} and train a diffusion process 
for brain lesion detection, see Fig.~\ref{fig:framework}-b.
In a classical DPM-based brain lesion detection \cite{rahman2023ambiguous}, only the ground-truth segmentation mask $\mathbf{x}_b$ 
and the original images
are required during the training phase; no additional information is provided throughout the entire diffusion process.
In contrast, for a given image $\mathbf{b}$
from the data pair $(\mathbf{b},\mathbf{x}_b)$ in the dataset $\mathcal{D}$, our objective is to generate a meaningful segmentation mask $\mathbf{x}_{b,0}$ similar to $\mathbf{x}_b$ during the diffusion process, incorporating the inter-discrepancy features $\mathcal{X}$ and intra-discrepancy features $\mathcal{Y}$ generated from the pre-trained AE models for modules \text{AE-1} and \text{AE-2}, as illustrated in Section \ref{Discrepanvy_feature_Generation}.

To achieve the objective, we induce anatomical information by concatenating those features as prior information of $\mathbf{x}_b$, thereby defining $\mathbf{X} := \mathbf{b}\oplus \mathcal{X} \oplus \mathcal{Y} \oplus \mathbf{x}_b$, where $\oplus$ is the concatenation operator.
During the forward diffusion process, the noise is only added to the ground-truth segmentation mask $\mathbf{x}_b$ at each step $t$ sampled from $\{1, \ldots, T\}$, i.e.,
\begin{equation}
\mathbf{x}_{b,t}=\sqrt{\overline{\alpha}_t}\mathbf{x}_b+\sqrt{1-\overline{\alpha}_t}\epsilon,\quad\mathrm{~with~}\epsilon\sim\mathcal{N}(0,\mathbf{I}).
\label{eq:add_noise_mask}
\end{equation}
The prior information from the concatenation is subsequently denoted as 
\begin{equation} \label{eqn:x_t-concate}
\mathbf{X}_t := \mathbf{b}\oplus \mathcal{X} \oplus \mathcal{Y} \oplus \mathbf{x}_{b,t}.
\end{equation}
The noise model $\epsilon_\theta$ is trained to approximate the ground-truth noise 
$\epsilon$ by being updated through gradient descent on the new objective function 
\begin{equation} L_{\text{simple}}=E_{t,\mathbf{x}_0,\epsilon}\left[\left\|\epsilon-\epsilon_\theta\left(\mathbf{X}_t,t\right)\right\|^2\right].
\label{eq:original_loss_function}
\end{equation}

\begin{algorithm}[h]
\KwIn{Brain image dataset $\mathcal{D}$}
\KwOut{Noise model $\epsilon_\theta$}
\nl {\bf repeat} \\
\nl \qquad $(\mathbf{b},\mathbf{x}_b) \sim \mathcal{D}$ \\
\nl \qquad  $\mathcal{X, Y} \leftarrow \mathbf{b}$ \\
\nl \qquad $t \sim {\tt Uniform}(\{1, \ldots, T\})$ \\
\nl \qquad $\epsilon \sim \mathcal{N}(0, \mathbf{I})$ \\
\nl \qquad $\mathbf{x}_{b,t}=\sqrt{\overline{\alpha}_t}\mathbf{x}_b+\sqrt{1-\overline{\alpha}_t}\epsilon$  \quad (i.e. Eq. \eqref{eq:add_noise_mask})\\
\nl \qquad $\mathbf{X}_t \leftarrow \mathbf{b}\oplus \mathcal{X} \oplus \mathcal{Y} \oplus \mathbf{x}_{b,t}$ \quad (i.e., Eq. \eqref{eqn:x_t-concate}) \\
\nl \qquad Update $\epsilon_\theta$ using the loss in Eq. \eqref{eq:original_loss_function} \\
\nl {\bf until} converged
\caption{{\bf The training process of {DDMD}} 
\label{Algorithm:1}}
\end{algorithm}

\begin{algorithm}[h]
\KwIn{Original brain MRI image $\mathbf{b}$ }
\KwOut{The predicted mask $\mathbf{x}_{b,0}$}
\nl Sample $\mathbf{x}_{b,T} \sim \mathcal{N}(0, \mathbf{I})$\\
\nl $\mathcal{X, Y} \leftarrow \mathbf{b}$ \\
\nl {\bf for } $t$ from $T$ {\bf to} $1$ {\bf do} \\
\nl \qquad $\mathbf{X}_t \leftarrow \mathbf{b}\oplus \mathcal{X} \oplus \mathcal{Y} \oplus \mathbf{x}_{b,t}$ \quad (i.e., Eq. \eqref{eqn:x_t-concate})  \\
\nl \qquad $\mathbf{z}\sim\mathcal{N}(0,\mathbf{I})$ \\  
\nl \qquad $\mathbf{x}_{b,t-1}=\frac{1}{\sqrt{\alpha_{t}}}\big(\mathbf{x}_{b,t} \!-\! \frac{1-\alpha_{t}}{\sqrt{1-\overline{\alpha}_{t}}}\epsilon_{\theta}(\mathbf{X}_{t},t)\big)+\sigma_{t}\mathbf{z}$  \\
\nl {\bf end} 
\caption{{\bf The sampling process of {DDMD}} 
\label{Algorithm:2}}
\end{algorithm}

For the reverse diffusion process, the predicted mask $\mathbf{x}_{b,0}$ can be predicted from $\mathbf{x}_{b,T}$ iteratively by
\begin{equation}
     \mathbf{x}_{b,t-1}=\frac{1}{\sqrt{\alpha_{t}}}\left(\mathbf{x}_{b,t}-\frac{1-\alpha_{t}}{\sqrt{1-\overline{\alpha}_{t}}}\epsilon_{\theta}(\mathbf{X}_{t},t)\right)+\sigma_{t}\mathbf{z},
\label{eq:inference result_DDMD}
\end{equation} 
with $\mathbf{z}\sim\mathcal{N}(0,\mathbf{I})$ and $t=T,...,1$ ({\it cf.} Eq. \eqref{eq:inference result}). 
The whole training and sampling processes of the our DDMD are summarized in Algorithms \ref{Algorithm:1} and \ref{Algorithm:2}, respectively.

\begin{table*}[b]
% \captionsetup{width=0.7\linewidth}
\caption{Performance comparison of different methods under different metrics on the BRATS2020 dataset. Note that
$\mathcal{X}$ and $\mathcal{Y}$ denote the inter-discrepancy features and the intra-discrepancy features, respectively. For all metrics, the mean ± standard deviation across five different folds is reported. The best results are highlighted in bold.}
\begin{tabularx}{0.7\linewidth}{l c r *{3}{>{\centering\arraybackslash}X}}
\hline
\multicolumn{3}{l}{Model}     & Dice [\%] & Miou [\%] & PA [\%] \\ \hline \hline
\multicolumn{3}{l}{\textit{Unet} \cite{ronneberger2015u}}      
& 74.16$\pm$1.29 & 82.50$\pm$1.14 & 85.54$\pm$1.45           \\
\multicolumn{3}{l}{\textit{SegNet} \cite{badrinarayanan2017segnet}}  
& 75.08$\pm$2.32 & 82.90$\pm$0.74 & 86.10$\pm$1.65           \\
\multicolumn{3}{l}{{\textit{DeepLabv3+}} \cite{chen2018encoder}} 
& 78.92$\pm$1.46 & 85.02$\pm$1.33 & 89.24$\pm$2.14         \\
\multicolumn{3}{l}{{\textit{CIMD}} \cite{rahman2023ambiguous}}   
& 81.15$\pm$1.25 & 86.61$\pm$1.19 & 89.68$\pm$1.12          \\
\hhline{------} % Dashed line for the second set of models
 \qquad \qquad   &  $\mathcal{X}$ & $\mathcal{Y}$ & \multicolumn{1}{l}{} &      &             \\
\hhline{------}
\multirow{3}{*}{\textit{Our DDMD}} 
& \usym{2717}   & \usym{2717}   
& 79.25$\pm$0.85   & 85.15$\pm$1.13  & 90.50$\pm$1.12     \\
& \usym{2713}  & \usym{2717}   
& \textbf{84.26$\pm$0.83}   & \textbf{88.58$\pm$1.22}  & \textbf{92.32$\pm$0.81}    \\
& \usym{2713}  & \usym{2713}   
& 74.63$\pm$1.30   & 82.81$\pm$0.65  & 84.97$\pm$1.36     \\ \hline
\end{tabularx}
\label{table:1}
\end{table*}

\section{Experimental Setup}\label{sec:Experimental Setup}
\subsection{Data}
We evaluate our DDMD method and its comparison with state-of-the-art methods using the publicly available BRATS2020 dataset \cite{menze2014multimodal}. The dataset provides four distinct MRI sequences, namely T1-weighted (T1), T2-weighted (T2), FLAIR, and T1-weighted with contrast enhancement (T1ce), including the corresponding pixel-wise ground-truth segmentation masks for each patient.
As we focus on 2D segmentation and the occurrence of tumours in the middle of the brain, we exclusively consider the middle axial slices for each patient. There are four MR sequences, which are stacked to a size of $(4, 256, 256)$, corresponding to each slice, with the corresponding ground-truth mask size being $(256, 256)$.
The image-level labels are determined based on the presence or absence of tumours in every ground-truth segmentation mask. 
To ensure a balanced data distribution, the training set consists of 5,643 instances of healthy slices and an equivalent number of unhealthy slices.
The generation of discrepancy features involves splitting the training data into datasets $\mathcal{I}_A$ and $\mathcal{I}_B$. In particular, dataset $\mathcal{I}_A$ contains all the healthy data, while dataset $\mathcal{I}_B$ encompasses the entire training set. The test set comprises 1,150 slices with tumours and 660 slices without tumours for evaluation.

\subsection{Evaluation Metrics}
To assess the effectiveness of brain lesion detection performance, several metrics are employed, such as dice coefficient (Dice), mean intersection over union (Miou), and pixel accuracy (PA). PA is calculated as the ratio of the number of correctly classified pixels against the total number of pixels. Miou is the ratio of the area of intersection between the predicted segmentation mask and the ground-truth segmentation mask against the area of the union between the two masks. Dice is the ratio of twice the area of intersection between the predicted segmentation mask and the ground-truth segmentation mask against the sum of the areas of the two masks. These metrics play a crucial role in providing a multifaceted assessment, addressing spatial overlap, agreement extent, and the overall pixel-wise accuracy especially in medical imaging. Higher values in these metrics reflect superior performance in the realm of medical image segmentation.

\subsection{Implementation Details}
The proposed method is implemented using the PyTorch framework. We utilize an AE structure including both an encoder and a decoder to generate discrepancy features through reconstruction. Specifically, the encoder is composed of four convolutional layers, while the decoder consists of four deconvolutional layers. The connection between the encoder and decoder involves three fully connected layers. In AE-1 and AE-2, the parameter $L$ is set to $3$. Each model is trained for $200$ epochs, employing the MSE loss and Adam optimizer with a fixed learning rate of $1 \times 10^{-4}$.
For the diffusion process, our proposed method utilizes a linear noise schedule with a time step set as $T = 1,000$ for all experiments. We also use the Adam optimizer to optimize the diffusion model, with a learning rate fixed as $1 \times10^{-4}$ and a batch size of $10$. The rest of the parameters for diffusion in the model are the same as that in \cite{nichol2021improved}. We train the model for $120,000$ iterations on an NVIDIA RTX 3090, which takes around 36 hours.

\subsection{Comparisons with Other Methods}
We compare our proposed method with current state-of-the-art approaches, including \textit{Unet} \cite{ronneberger2015u}, \textit{SegNet} \cite{badrinarayanan2017segnet} and \textit{DeepLabv3+} \cite{chen2018encoder}, specifically designed for multi-modal segmentation tasks. All these three models employ an Adam optimizer with a learning rate of $1 \times10^{-5}$ and a batch size of $32$ during the training process. 
The training is performed using the negative log-likelihood loss function, which stops after 50 epochs. 
Additionally, our method is also compared with a diffusion-based lesion detection method known as {collectively intelligent medical diffusion (CIMD)} \cite{rahman2023ambiguous}.
We adopt all hyperparameter configurations recommended in the original implementation as detailed in \cite{rahman2023ambiguous}.

To evaluate the effects of inter-discrepancy and intra-discrepancy on the generated segmentation mask in response to the original image input, we utilize three distinct strategies within our framework {DDMD} as part of an ablation study.
Depending on how features are combined in the diffusion process ({\it cf.} Eq. \eqref{eqn:x_t-concate}), these strategies can be categorized into configurations comprising: i) solely the original brain modalities, i.e., {DDMD} w/o $\mathcal{X}$\&$\mathcal{Y}$, named {\it DDMD-mini}; ii) brain modalities with inter-discrepancy features, i.e., {DDMD} w/ $\mathcal{X}$ \& w/o $\mathcal{Y}$, named {\it DDMD-light}; and iii) brain modalities integrated with both inter-discrepancy and intra-discrepancy features, i.e., {DDMD}.

\section{Results and Discussion}\label{sec:Results and Discussion}
% \subsection{Brain Lesion Detection}

\subsection{Quantitative Comparisons}

\begin{figure*}[tb]
% \vspace{0.2cm}
\centering
\includegraphics[width=0.97\textwidth, height=0.72\textheight]{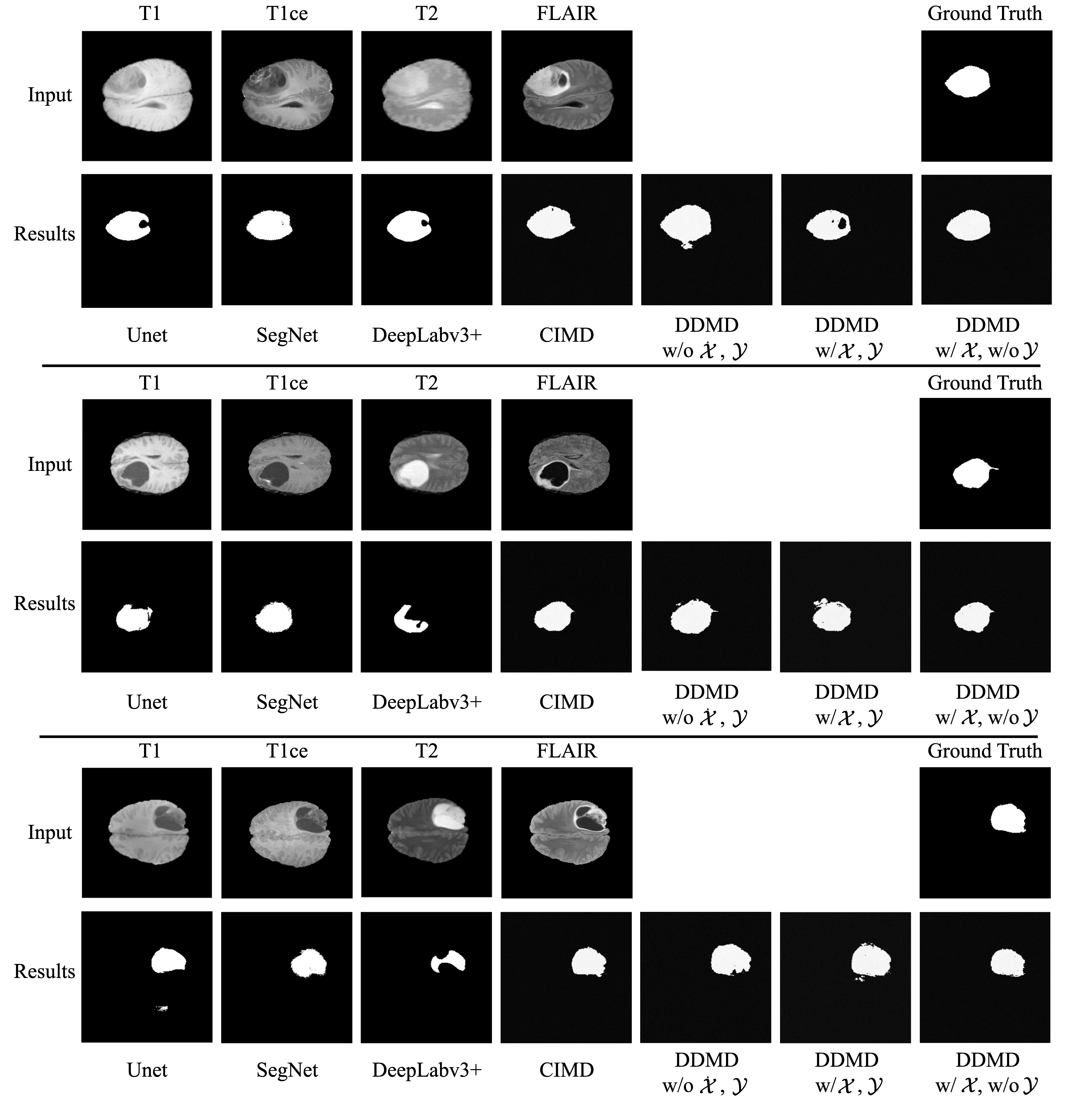}
\caption{Qualitative comparison between different brain lesion detection methods. Three examples are showcased here. For each example, the first row illustrates the input brain images at four different modalities (i.e., T1, T1ce, T2 and FLAIR) along with the ground-truth segmentation mask, and the second row displays the segmentation mask achieved by each method.}
\label{fig:all_results2}
% \vspace{-0.5cm}
\end{figure*}

\begin{figure*}[b]
% \vspace{-0.2cm}
\centering
\includegraphics[width=0.98\textwidth, height=0.50\textheight]{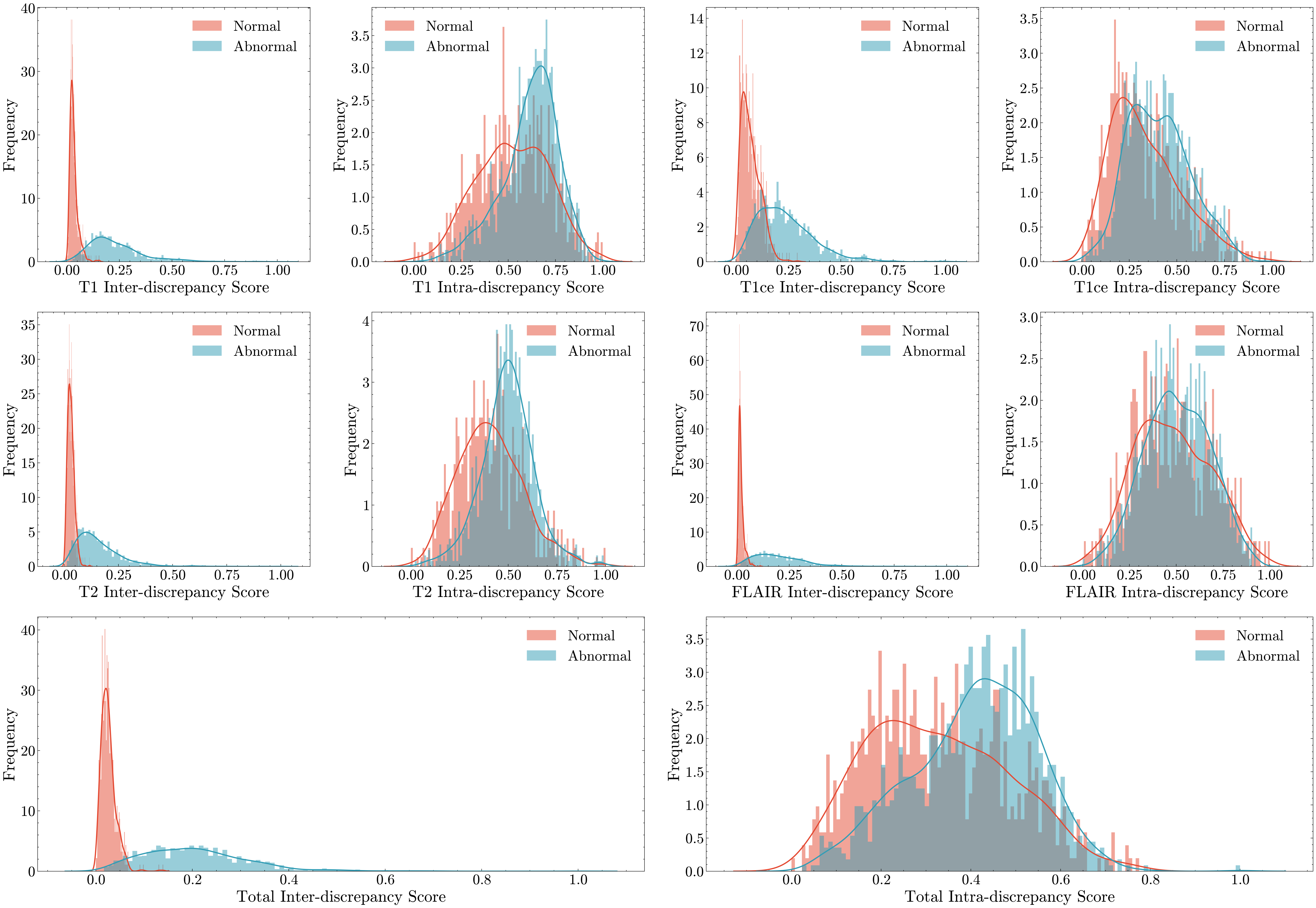}
\caption{Histograms of the inter-discrepancy and intra-discrepancy scores obtained from the normal and abnormal images in the test dataset for different data modalities. The first and second rows show the results across the individual data modalities (i.e., T1, T1ce, T2 and FlAIR), and the third row displays the results under the whole data modalities. Scores are normalized to [0, 1].}
\label{fig:distribution_results}
% \vspace{-0.5cm}
\end{figure*}
Table \ref{table:1} provides a quantitative comparison between different methods on brain lesion detection in terms of metrics Dice, Miou and PA. For our {DDMD} and its variations, we take the original images $\mathbf{b}$ from the test set and produce the segmentation mask following Algorithm \ref{Algorithm:2}. Five iterations of the algorithm is conducted to sample five distinct segmentation masks for each image. The final segmentation mask is determined by averaging these five masks and applying a threshold of 0.5.
It can be observed from Table \ref{table:1} that our approach ({DDMD}-light) outperforms other evaluated methods under all the metrics. Compared with the end-to-end segmentation models (i.e., {Unet} \cite{ronneberger2015u}, {SegNet} \cite{badrinarayanan2017segnet} and {DeepLabv3+} \cite{chen2018encoder}), the DPM-based models (e.g. {CIMD} \cite{rahman2023ambiguous} and our {DDMD}-mini) show significant advantages on brain lesion detection tasks.

Fig. \ref{fig:all_results2} provides a visual comparison of some predicted segmentation results obtained from various brain lesion detection methods. Each example in Fig. \ref{fig:all_results2} showcases the original input brain images in four different modalities (i.e., T1, T1ce, T2, and FLAIR) along with their corresponding ground-truth segmentation mask and the predicted segmentation masks by different methods (with the threshold set to 0.5 for binary mask generation).
Unlike end-to-end models that rely on convolutional neural networks and may suffer from overfitting and limited receptive fields, DPM models utilize probabilistic modelling and diffusion processes to effectively capture long-range dependencies and contextual information. This enables DPM-based models to handle complex brain structures with greater accuracy and robustness, especially when subtle boundaries or intricate anatomical details are critical, as shown in Fig \ref{fig:all_results2}. 

It is notable that {CIMD} employs its model architecture to regulate the diversity of the sampling process during training. In comparison, our {DDMD}-mini solely relies on the original image information, leading to slight performance.
The performance of {DDMD} in integrating inter-discrepancy and intra-discrepancy features shows notable differences. Interestingly, when compared to {DDMD}, which combines both the inter-discrepancy and intra-discrepancy features, {DDMD}-light, which concentrates only on the inter-discrepancy features, exhibits better performance. A comprehensive analysis of this observation is provided in the subsequent section.

\begin{figure*}[b]
% \vspace{-0.2cm}
% \label{fig:framework}
\centering
\includegraphics[width=0.98\textwidth, height=0.56\textheight]{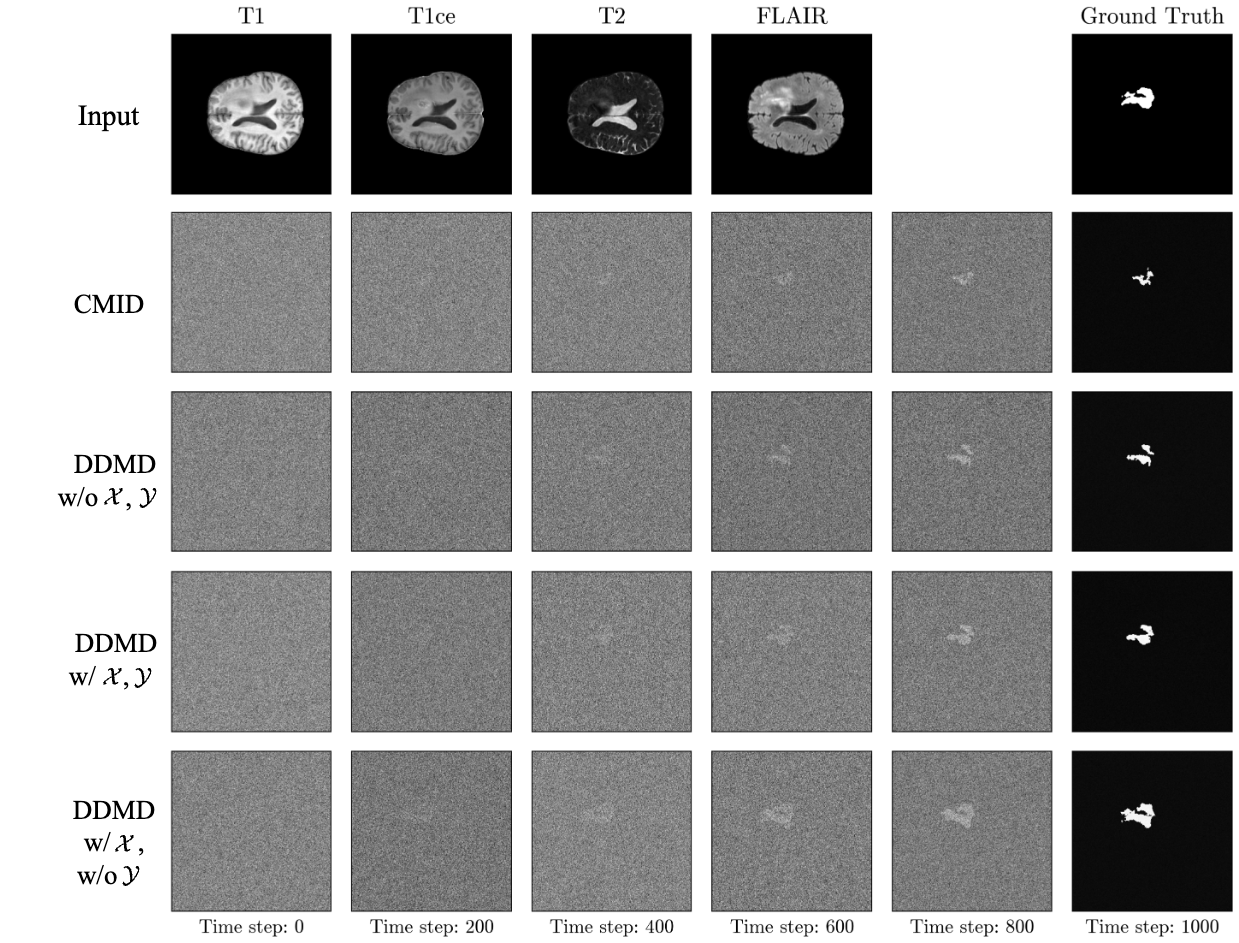}
\caption{Diffusion results of different sampling methods at different time steps. The first row shows the input MRI images with T1, T1ce, T2 and FLair modalities and the corresponding ground-truth segmentation mask. The second to fifth rows present intermediate results of the {CIMD}, {DDMD}-mini, {DDMD}, and {DDMD}-light sampling processes, respectively.}
\label{fig:diffusion_results}
% \vspace{-0.5cm}
\end{figure*}
\subsection{Ablation Study}\label{sec:ablation_study}

From Table \ref{table:1}, we see that combining the original images with both the inter-discrepancy features and intra-discrepancy features does not lead to a substantial enhancement in the segmentation results in comparison to the way of only incorporating the original images. Conversely, significant improvement in segmentation outcomes is achieved when the original images are exclusively combined with the inter-discrepancy features. It is important to note that, the inter-discrepancy features pertain to the inter-class difference, signifying distinctions between normal and abnormal data, while intra-discrepancy features represent intra-class differences, denoting variances within the same category. 

As mentioned in Section \ref{Discrepanvy_feature_Generation}, the discrepancy features are generated from the pre-trained reconstruction modules $\text{AE-1}$ and $\text{AE-2}$ and both the inter-discrepancy features $\mathcal{X}$ and intra-discrepancy features $\mathcal{Y}$ are calculated from all data modalities (i.e., T1, T1ce, T2, and FLAIR), with the same size  $H \times W$. 
Similarly, based on the reconstruction results from the AE-1 and AE-2 modules, we can also extract the features corresponding to each data modality of every test image from the corresponding channel in the reconstruction results, and calculate the inter-discrepancy score and intra-discrepancy score for each data modality of every test image (see Section \ref{Discrepanvy_feature_Generation}).
Fig. \ref{fig:distribution_results} presents the histograms of the discrepancy scores obtained from the normal and abnormal images in the test dataset for different data modalities, where the x-axis and y-axis of each plot denote the score and the frequency, respectively.
In Fig. \ref{fig:distribution_results}, it is evident that the overlap of the histograms of the inter-discrepancy scores is notably lower compared to that of the intra-discrepancy scores across all the data modalities. This implies a strong discriminative capacity difference between these two types of discrepancy features in distinguishing normal and abnormal images.
Therefore, in contrast to the feature representation provided by intra-discrepancy features, inter-discrepancy features are notably to be more effective in the generation of pixel-level annotations derived from image-level annotations.

\subsection{Further Diffusion Results}
Intermediate results from method {CIMD} and our {DDMD} and its variants in the reverse diffusion process are compared in Fig. \ref{fig:diffusion_results}. 
Throughout the iterative steps, it is evident that each method initiates from the Gaussian noise input at time step 0 and systematically produces the corresponding segmentation masks from the initial noise based on the respective input conditions. {CIMD} and {DDMD}-mini, which generate segmentation masks solely from the original brain data modalities (i.e., T1, T1ce, T2, FLAIR), exhibit distinctions from the ground-truth segmentation masks (see the second and third rows in Fig. \ref{fig:diffusion_results}). In comparison to {DDMD}, the results of {DDMD}-light, which only incorporates the inter-discrepancy features, demonstrate closer alignment with the ground truth (see the forth and fifth rows in Fig. \ref{fig:diffusion_results}).

Remarkably, as the reverse diffusion process goes on, notable enhancements in image quality become apparent, characterized by progressively reduced noise levels, leading to increasingly realistic segmentation results. At each step of the reverse diffusion process, the DPM model consistently references the original input conditions, employing probabilistic modelling and diffusion processes to effectively capture long-range dependencies and contextual information. This aspect further underscores the efficacy of the DPM model in brain lesion detection compared to end-to-end models.

\section{Conclusion}\label{sec:conclusion}

In this work, we proposed a DPM-based lesion detection network {DDMD}, which can generate brain lesion segmentation masks for given images. 
In contrast to existing alternative approaches, {DDMD} and its variants leverage the discrepancy features instead of relying directly on image-level annotations or the original brain modalities. The inconsistencies obtained from the image-level annotations are converted into distributional disparities among diverse samples while preserving the information within homogeneous samples. Thorough experiments and comparisons demonstrated the great performance of our methods in brain lesion segmentation masks prediction. Moreover, our methods can be incorporated into any diffusion-based framework with minimal additional training. Future research avenues may explore the extension of {DDMD} to broader computer vision problems and its applicability to diverse medical imaging modalities. It might also be interesting to explore the extension of {DDMD} in investigating more invariant features in pixel-level annotations.

\section*{Acknowledgements}
MN's contribution to this work was funded by Grant EP/S000356/1, Artificial and Augmented Intelligence for Automated Scientific Discovery, Engineering and Physical Sciences Research Council (EPSRC), UK.

\printcredits

% \bibliographystyle{cas-model2-names}
\bibliographystyle{IEEEtran}
% \bibliographystyle{unsrt}
\bibliography{cas-refs}

\end{document}